\documentclass{article}
\usepackage{M-PSI}
\usepackage{amsmath,amssymb,amsfonts}
\usepackage{algorithmic}
\usepackage{graphicx}
\usepackage{textcomp}
\usepackage{xcolor}
\usepackage{url}
\usepackage{multirow}
\usepackage{enumitem}
\def\BibTeX{{\rm B\kern-.05em{\sc i\kern-.025em b}\kern-.08em
    T\kern-.1667em\lower.7ex\hbox{E}\kern-.125emX}}

\usepackage[switch]{lineno}

\author{Xi Chen\textsuperscript{1,2}, Julien Cumin\textsuperscript{1}, Fano Ramparany\textsuperscript{1}, \href{https://research.vaufreydaz.org/}{Dominique~Vaufreydaz\textsuperscript{2,~\small{\ExternalLink }}}\vspace{0.1cm}\\
{$^1$ Orange Innovation}\\
{$^2$ Univ. Grenoble Alpes, CNRS, Grenoble INP, LIG, 38000 Grenoble, France}\\ % M-PSI Affiliation
}

\title{Generative Resident Separation and Multi-label Classification for Multi-person Activity Recognition}

\usepackage[pdfencoding=auto, pdfusetitle]{hyperref}
\hypersetup{
  hidelinks,
  urlcolor=red,
  pdftitle={Generative Resident Separation and Multi-label Classification for Multi-person Activity Recognition},
  pdfauthor={Xi Chen, Julien Cumin, Fano Ramparany, Dominique Vaufreydaz},
  pdfkeywords={Human Activity Recognition, Large Language Models, Ambient Intelligence, Smart Homes.},
}

\author{Xi Chen\textsuperscript{1,2}, Julien Cumin\textsuperscript{1}, Fano Ramparany\textsuperscript{1}, \href{https://research.vaufreydaz.org/}{Dominique~Vaufreydaz\textsuperscript{2,~\small{\ExternalLink }}}\vspace{0.1cm}\\
{$^1$ Orange Innovation}\\
{$^2$ Univ. Grenoble Alpes, CNRS, Grenoble INP, LIG, 38000 Grenoble, France}\\ % M-PSI Affiliation
}

\removefirstpagelogo{}
\begin{document}
% \linenumbers % Added for review (Doms)

\begin{abstract}
This paper presents two models to address the problem of multi-person activity recognition using ambient sensors in a home. The first model, Seq2Res, uses a sequence generation approach to separate sensor events from different residents. The second model, BiGRU+Q2L, uses a Query2Label multi-label classifier to predict multiple activities simultaneously. Performances of these models are compared to a state-of-the-art model in different experimental scenarios, using a state-of-the-art dataset of two residents in a home instrumented with ambient sensors. These results lead to a discussion on the advantages and drawbacks of resident separation and multi-label classification for multi-person activity recognition.
\end{abstract}

\section{Introduction}
% 1. 总体介绍任务，应用
% 2. 难点
% 3. 贡献
% 4. organization
Ambient-based activity recognition has garnered growing interest due to its non-intrusive, privacy-friendly, and cost-effective properties. This technology leverages ambient sensors strategically placed in the environment (such as a home) to capture changes and interactions in their proximity. These recorded changes are referred to as sensor events. By analysing sequences formed by these sensor events, residents' activities in the environment can be identified. However, in real-home scenarios, there are often multiple residents, and sensor events captured in these situations correspond to potentially multiple and intertwined activities. Activity recognition in such situations is referred to as \textbf{multi-person activity recognition}.

The primary challenge in multi-person activity recognition is to separate activity information for each person. Existing methods can generally be categorized into 2 classes based on when this separation occurs: \textbf{resident separation} and \textbf{multi-label classification}. On one hand, resident separation aims to distinguish sensor events triggered by different residents, and subsequently, perform individual activity recognition on each separated sensor event sequence. On the other hand, multi-label classification methods involve extracting global features from sensor event sequences and then using these features to recognize multiple activity classes associated with individuals, thereby recognizing multi-person activities.

This paper presents several approaches: one based on resident separation, called Seq2Res, and another based on multi-label classification, called BiGRU+Q2L. A third approach combines them into a two-stage model. Unlike previous separation approaches that assign sensor events to residents one by one, Seq2Res employs a Sequence-to-Sequence (Seq2Seq)~\cite{sutskever2014sequence} architecture. It models the entire sensor sequence and generates separated sequences based on the modeled context. On the other hand, BiGRU+Q2L uses attention mechanisms to establish correlations not only among activity labels but also between labels and features. This enables a more accurate and flexible multi-label classification. Finally, the two approaches are combined in a model that separates resident information while considering the correlation of residents' activities.

This paper is organized as follows: Section~\ref{section:sota} provides a summary of related work on resident separation and multi-label classification. Section~\ref{section:methods} presents the Seq2Res and BiGRU+Q2L models, as well as their combination in a two-stage model. Section~\ref{section:exps} describes the experimental results of these models as well as state-of-the-art models, on a state-of-the-art dataset. Finally, a conclusion is given in Section~\ref{section:conclusion}.

\section{Related Work}
\label{section:sota}
% 目前的基于环境传感器的多住户活动识别方法可以根据进行分离的层级分为居民分离和多标签分类。居民分离旨在将原始数据在输入端分配到每个居民之后，再对每个居民的专属数据进行单独的活动识别。而多标签分类方法对原始数据进行整体的特征提取后，在分类端将特征分配到多个活动类别中，从而识别出多人活动。本节针对这两种类别介绍已知的工作。
% The primary challenge in multi-person activity recognition is to separate activity information for each person. Existing methods can generally be categorized into 2 classes based on when this separation occurs: resident separation and multi-label classification. Resident separation aims to assign parts of collected data to each person, and then perform individual activity recognition on this individual data for each person. On the other hand, multi-label classification methods involve extracting global features from all data, and then use these features to recognize multiple activity classes, thereby recognizing multi-person activities.

%This section discusses existing works related to both approaches.

\subsection{Resident Separation}
Crandall and Cook~\cite{crandall2008attributing} use a supervised Naïve Bayes model to assign each sensor event to specific residents of a home, a problem often called \textbf{data association} in the literature. This method is highly reliant on the timing of events and resident schedule habits for classification, without consideration for spatiotemporal relationships between sensor events. Riboni \textit{et al.}~\cite{riboni2020unsupervised} modeled these spatiotemporal relationships by conducting a statistical analysis of the co-occurrence frequency of two adjacent sensor events within a defined temporal window in single-person data. If two sensor events, with rare co-occurrences in single-person data, happen in multi-person data within the defined temporal window, it suggests they come from different residents. In this approach, training requires pre-separated single-person data. Arrotta \textit{et al.}~\cite{arrotta2023micar} presented MICAR, a knowledge-based approach for data association, where sensor events are assigned to corresponding residents using ontological reasoning on context. While less affected by data scarcity, it relies on explicit information like the location of each user, which may not always be available.

Bouchabou \textit{et al.}~\cite{bouchabou2021fully} drew inspiration from language models used in natural language processing. Each sensor event is considered as a token and modeled using a word embedding skip-gram model~\cite{mikolov2013distributed}. The advantage of this approach is its ability to establish more flexible and richer event correlations and can be directly applied in multi-person data. Similarly, SMRT ~\cite{wang2020smrt} and GAMUT ~\cite{wang2021multi} also adopted skip-gram models to map sensors into a latent space. In addition, these methods used a linear Gaussian dynamic model and a Gaussian Mixture Probability Hypothesis Density~(GM-PHD) filter to track residents' states in the latent space. 
While these probabilistic models enable unsupervised resident separation, their use of the linear dynamic model results in uneven tracking capabilities for residents or pets with diverse mobility profiles, rendering the model sensitive to anomalous sensor events.

\subsection{Multi-label classification}
%A significant amount of work didn't directly seek resident separation at the sensor event level but rather modeled the entire sequence and classified it into multiple classes. Depending on the requirements, these classes can be user-specific, labeled with a combination of resident identifier and activity, or anonymous, solely representing the activity. Indeed, the features of the data are allocated to the corresponding classes during the classification stage, thereby achieving separation implicitly.

To identify the resident associated with a specific activity without resident separation, multi-label classification methods typically combine each activity class with a corresponding resident identifier. Alternatively, these methods may opt for anonymous activity classification, without predicting the individual responsible for each activity.

The most straightforward multi-label classification method is binary relevance~\cite{kumar2015multi,alhamoud2016activity,jethanandani2019binary}, where each activity label is predicted by an individual binary classifier. The problem of this approach is its inability to consider the interdependence between activity classes. For instance, in real-life scenarios, activities like ``\textit{User A using the toilet}'' and ``\textit{User B using the toilet}'' are less likely to occur simultaneously. Binary relevance struggles to learn such patterns because predictions for two labels are independent. Extending binary relevance, the classifier chain method~\cite{mohamed2017multi,jethanandani2020multi} introduces dependencies between binary classifiers by using the output of one classifier as a feature for the next classifier. However, the performances of such methods are highly dependent on the ordering of classifiers. Another extension of the binary relevance method is the work of Liu \textit{et al.}, who propose Query2Label~(Q2L)~\cite{liu2021query2label}. This method embedded labels as vectors and used Transformer decoders \cite{vaswani2017attention} to model inter-label relationships and then queried the label-related features from the feature space.

A number of works used a label combination method~\cite{mohamed2017modeling, benmansour2017modeling, chen2022transformer}. This method defines combinations of activities that are performed simultaneously by different persons as new labels and uses single-label classifiers to predict these combinations. As such, dependencies between the initial activity labels are hard-coded as the new labels, reducing the training complexity and often resulting in higher performances. For example, Chen \textit{et al.}~\cite{chen2022transformer} achieve state-of-the-art performance with TransBiGRU, a combination of Transformer \cite{vaswani2017attention} and Bidirectional Gated Recurrent Units (BiGRU) for feature extraction, with a label combination classifier at the end. Label combination has three main drawbacks: complexity grows exponentially with the number of persons and activity classes; class imbalance is exacerbated; the trained model cannot predict combinations of classes that do not exist in the training set.

% 我们认为，现有的resident separation工作

\section{Proposed models}
\label{section:methods}

%In the following, we propose our novel methods for both resident separation and multi-label classification as well as integrate them as a two-stage model.

\subsection{Seq2Res resident separation model}
% 传感器事件和居民在之前的工作中主要是一对一的，对于多对一的能力较差。
% 居民的动态模型是线性的
% 超参数过多
% GAMUT和SMRT假设了马尔可夫假设，只根据上一个状态预测下一个，无法建模更加复杂的运动，且在传感器密度高的地方难以分离。我们摒弃了一个个地将事件进行分离的方法，而是考虑更长期的整体建模和整体分离。这样的好处是能够考虑context，而且对噪音更加有鲁棒性，且生成的序列会更加流畅。在数据量允许的情况下，能够学习到更加复杂多变的情况。
% 进一步relaxes constraints of previous algorithms

% \begin{figure*}
% \centering
% \begin{minipage}[b]{.55\textwidth}
%     \includegraphics[width=1.0\textwidth]{Figures/Seq2Sep.png}
%     \caption{Framework of the proposed Seq2Res model for resident separation.}
%     \label{fig:Seq2Sep}
% \end{minipage}\hfill
% \begin{minipage}[b]{.43\textwidth}
%     \includegraphics[width=1.0\textwidth]{Figures/Q2L_classifier.png}
%     %\caption{The framework of the proposed BiGRU+Query2Label model. On one hand, a BiGRU extracts temporal features between input events. On the other hand, the Transformer decoder layer first models the correlations between labels with the self-attention module and then pools relevant features from the input feature space using cross-attention with the label vectors as queries. The pooled features for each label are fed into a binary classifier to predict the presence of that label.}
%     \caption{Framework of the proposed BiGRU+Q2L model, with feature extraction from sensor events depicted on the left, and Query2Label transformer decoder for multi-label classification on the right.}
%     \label{fig:Q2L}
% \end{minipage}
% \end{figure*}

\begin{figure}
    \centering
    \includegraphics[width=1.0\linewidth]{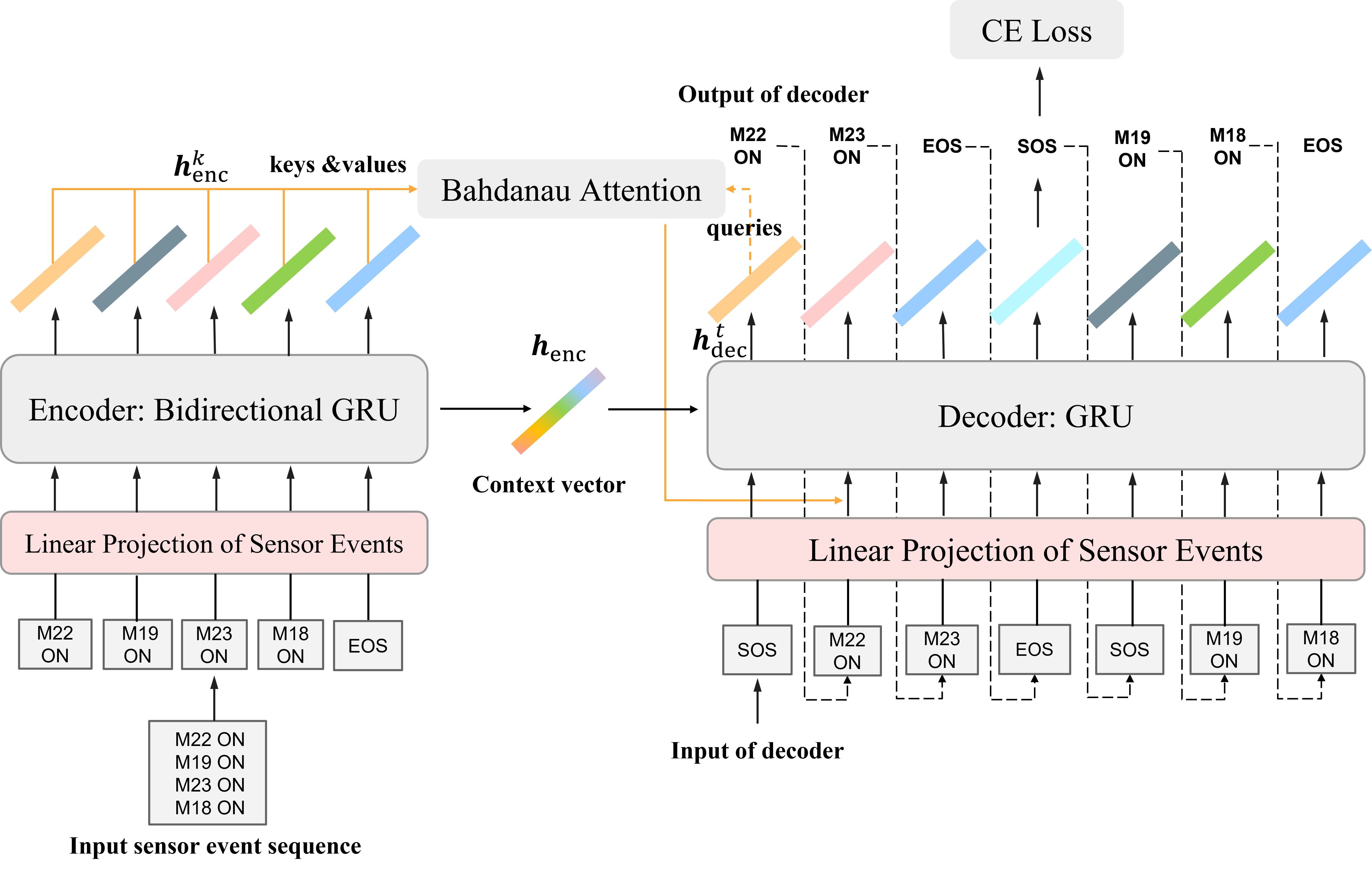}
    \caption{Framework of the proposed Seq2Res model for resident separation.}
    \label{fig:Seq2Sep}
\vspace{-1em}
\end{figure}

% Assumption, problem definition, objective
In this research, it is assumed that, as in \cite{riboni2020unsupervised, chen2022transformer}, there are two residents living in a smart home equipped with ambient sensors. Theoretically, this research can be extended to scenarios involving more than two residents, but this is left as future work. Given a sequence of sensor events $\{e_k\}$, the objective is to assign each $e_k$ to one of two sets $\{e_k^1\}$ and $\{e_k^2\}$ where each set represents an event sequence of a resident. Existing resident separation methods generally determine the belonging of the next sensor event based only on the resident's state at the previous time step. These methods not only overlook longer-term contextual information but also can lead to error accumulation. To address these issues, the present proposal attempts to relax the constraints of the previous algorithms. In contrast with previous approaches, a generative method is used to ``translate'' a sensor event sequence triggered by multiple individuals into separate event sequences for each resident. As a result, the separated sequences $\{e_k^1\}$ and $\{e_k^2\}$ are no longer constructed by partitioning sensor events one by one but are generated based on the overall context of the input sequence. This means that the two sequences no longer guarantee that $\{e^1_k\} \cup \{e^2_k\} = \{e_k\}$. %Differing from previous work, we are not seeking to allocate each event directly, but rather to collectively encode the sequence $\{e_k\}$ and then regenerate two separate partitions. 
An attention-based Sequence to Sequence (Seq2Seq) architecture is used, based on the work of Bahdanau \textit{et al.}~\cite{bahdanau2014neural}. Figure~\ref{fig:Seq2Sep} illustrates the proposed model, which is called \textbf{Seq2Res} (Sequence to Residents).

% The Seq2Seq architecture compresses and abstracts input sequences using an encoder model, representing them as a context vector. Subsequently, a decoder model utilizes this encoded vector to generate the target sequence. In this work, the encoder encodes the entire multi-person event sequence into a context vector, and the decoder generates separate event sequences for each resident based on this context vector. 

% Figure illustration and general explication of idea: input, encoder, decoder, output, label, loss

% input embedding, encoder-decoder, attention
\subsubsection{Input sequence encoding}
This step is illustrated on the left side of Figure \ref{fig:Seq2Sep}. As in~\cite{bouchabou2021fully}, a sensor event is represented as a numerical token. To better capture spatiotemporal relationships between sensors, the input token sequence is mapped into an embedding space of dimension $D$. Then, a bidirectional GRU is used to encode the bidirectional temporal characteristics of the event sequence, resulting in output vectors $\{\mathbf{h}_\text{enc}^k\}$ and a context vector $\mathbf{h}_\text{enc}$, with a dimension of $2D$. 

% Why set a label like this?
% 这种方法的关键问题在于，如何在生成一个的同时考虑到另一个的生成？最简单的方法是连续生成两个，这使得第一个的hidden state被第二个得知，从而在生成第二个的时候会考虑第一个生成过的东西。而第一个的生成由输入序列的第一个动作进行引导生成。第二个序列重头生成的问题在于，我们不知道第二个序列的开头，第二个序列的开头是要生成的，而他的开头在注意力上与输入序列的结尾没有距离上的联系。我们生成反转的第二个序列，使得EOS token与第二个序列的开头在隐变量上相似（因为距离接近）。此外，双向信息使得每个序列的生成只用到一个方向的信息，从而增加了两个序列的分离度。
\subsubsection{Output sequence generation}
% With the encoded context vector $\mathbf{h}_\text{enc}$ as the initial hidden state with a ``Start of Sequence'' token $<$SOS$>$ as input, decoder models can be used to generate separated sequences of sensor events for both residents. 
Designing decoders capable of generating separated sequences presents a significant challenge, particularly in generating one resident's event sequence while taking into account the generation of the other resident's sequence. A straightforward but effective approach involves employing a single decoder for the sequential generation of two sequences. This requires passing the hidden state from the first sequence generation to the subsequent sequence. In this work, the decoder is designed to produce a unified sequence where ${e_k^1}$ and ${e_k^2}$ are continuously generated. The initial hidden state for generating $\{e_k^2\}$ is thus the hidden state of $\{e_k^1\}$, and the initial hidden state for generating $\{e_k^1\}$ is $\mathbf{h}_\text{enc}$.

As illustrated in the right part of Figure~\ref{fig:Seq2Sep}, taking a ``Start of Sequence'' token SOS as input and $\mathbf{h}_\text{enc}$ as hidden state, a GRU-based decoder is applied to give an output vector and a new hidden state vector. The output is then mapped into the probability vector of events using a fully connected network and a softmax function. The event with the highest probability will be considered as the generated event in this step. Generated events will serve as input for the next step, prompting the decoder to generate based on the context and the existing sequence. After the first sequence is generated, the model is trained to generate an ``End of Sequence'' token EOS, followed by an SOS to prompt the generation of the second sequence until a second EOS is finally generated. 

% Following this idea, the initial hidden state for the first sequence generation is $\mathbf{h}_\text{enc}$, with an input of $<$SOS$>$, while the initial hidden state for the second sequence generation is the hidden state of the first sequence $\mathbf{h}_\text{dec}^\text{R1}$, with an input of $<$SOS$>$ as well. 

% In summary, the decoder is trained to generate the event sequence for the first resident based on the context vector provided by the encoder. It generates an ``End of Sequence'' token $<$EOS$>$ to indicate the end of the sequence for the first resident. Subsequently, it generates a second $<$SOS$>$ token to prompt the continuation of sequence generation for the next resident. Finally, the generation process concludes with the generation of a second $<$EOS$>$ token, signaling the end of the entire generation process.

Since the generation of the first sequence depends on the first event, we set the resident who triggers the first event in the input sequence as resident 1. To train the model, labels of separated sequence are of the form $\{\{e_{k}^{1}\}, \text{EOS}, \text{SOS}, \{e_{k}^{2}\}, \text{EOS}\}$, where $\{e_{k}^{i}\}$ is the event sequence of resident $i$. Cross Entropy Loss (CE Loss) is used as the loss function.

\subsubsection{Bahdanau Attention}
Due to the inherent limitations of encoding the entire input with a single context vector, an attention mechanism is added to enhance the decoder's aligning capability. This allows the decoder to focus on different parts of the encoder's output at each step of the decoding process.

Specifically, the Bahdanau attention mechanism~\cite{bahdanau2014neural} is applied. In each decoding step $t$, the decoder's hidden state at the previous time step $\mathbf{h}_{\text{dec}}^{t-1}$ functions as the query. The encoder's outputs $\{\mathbf{h}_{\text{enc}}^{k}\}$ serve as both keys and values. Following the computation of Bahdanau attention scores, a weighted average is computed across the encoder's output vectors $\{\mathbf{h}_{\text{enc}}^{k}\}$, yielding a context vector $\mathbf{c}_t$. This context vector $\mathbf{c}_t$ is then concatenated with the embedding vector of the decoder's input at time $t$ and subsequently fed into the GRU. The Bahdanau attention score $s_{t, k}$ between the query $\mathbf{h}_{\text{dec}}^{t-1}$ and the key ${\mathbf{h}_{\text{enc}}^{k}}$ is computed as
$$s_{t, k} = \mathbf{v}^\top \text{tanh}( \mathbf{W} \mathbf{h}_{\text{dec}}^{t-1} + \mathbf{U}\mathbf{h}_{\text{enc}}^{k} + \mathbf{b}) ,$$
where $\mathbf{v}$, $\mathbf{W}$, $\mathbf{U}$, and $\mathbf{b}$ are learnable parameters.

\subsection{BiGRU+Query2Label multi-label classification model}

% \todo{New organisation to match results section: }

% \todo{1) Feature extractor with BiGRU (mention that you can add a fully connected layer / binary relevance? to directly classify multi-resident activities - rename to BiGRU+FCN or something else?) }

% \todo{2) Query2Label for multi-resident activity recognition (mention that this model is called BiGRU+Q2L)}

\begin{figure}[t]
    \centering
    \includegraphics[scale=0.28]{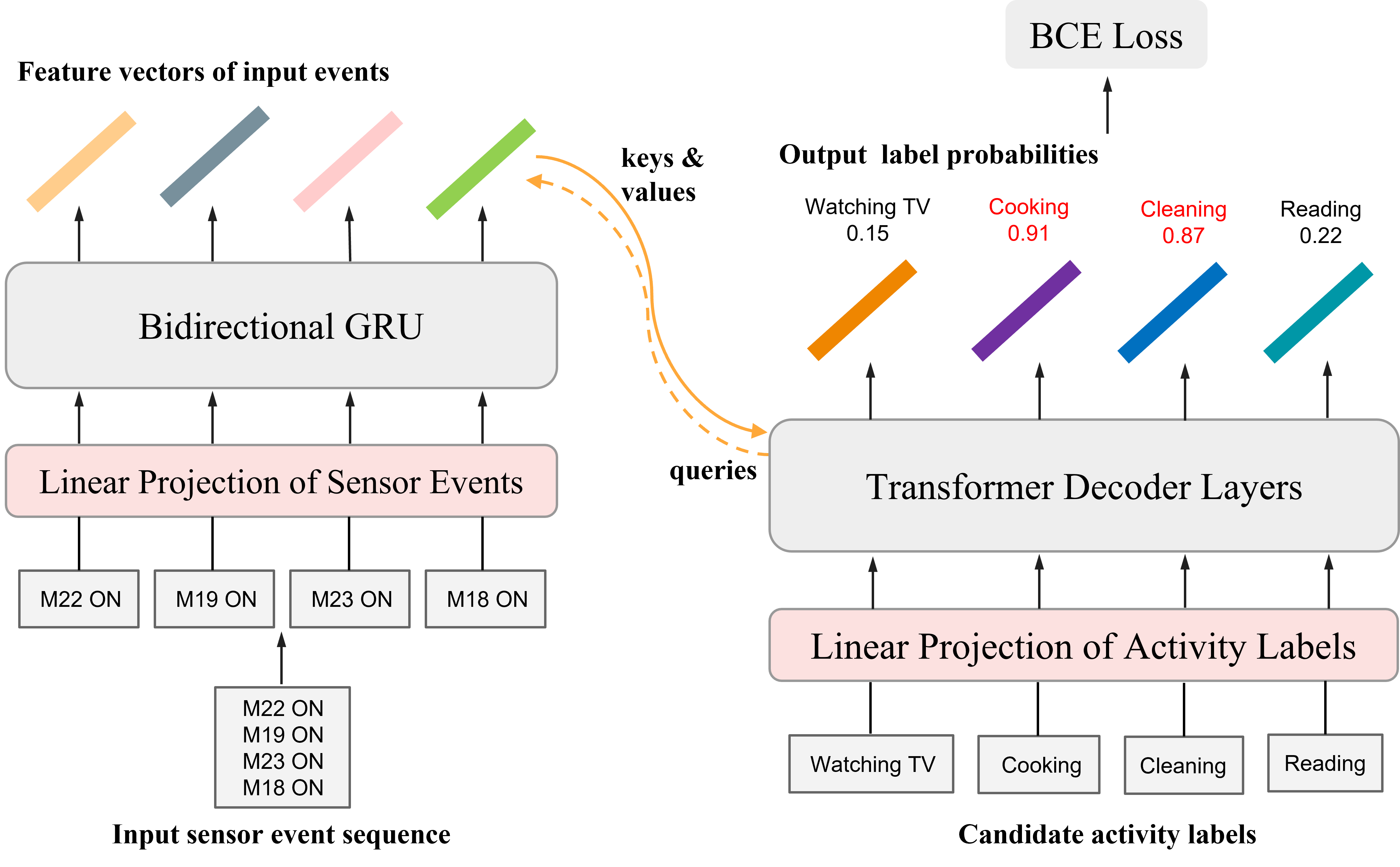}
    %\caption{The framework of the proposed BiGRU+Query2Label model. On one hand, a BiGRU extracts temporal features between input events. On the other hand, the Transformer decoder layer first models the correlations between labels with the self-attention module and then pools relevant features from the input feature space using cross-attention with the label vectors as queries. The pooled features for each label are fed into a binary classifier to predict the presence of that label.}
    \caption{Framework of the proposed BiGRU+Q2L model, with feature extraction from sensor events depicted on the left, and Query2Label transformer decoder for multi-label classification on the right.}
    \label{fig:Q2L}
\vspace{-1em}
\end{figure}

In the following, we introduce a multi-label classifier based on an attention mechanism, while employing a BiGRU model as a sequence feature extractor. Figure \ref{fig:Q2L} illustrates the framework of this method.

\subsubsection{BiGRU-based feature extractor}
The feature extractor of the proposed multi-label classification model is presented on the left side of Figure~\ref{fig:Q2L}. The input event sequence, whether separated or not, is linearly mapped to the embedding space and then processed by a BiGRU to extract bidirectional temporal information. In the TransBiGRU model~\cite{chen2022transformer}, 6 composite layers of Transformer encoder coupled with BiGRU are used. Following preliminary experiments on the same dataset as in \cite{chen2022transformer}, it appears that BiGRU is the important component, and Transformer layers play a lesser role. As such, only BiGRU layers are used for the encoder in this work. Comparative results between the two models can be found in Section \ref{sec: HAR_results}.

\subsubsection{Query2Label (Q2L) multi-label classifier}
\label{section:q2l}
Given an input event sequence $\{e_k\}$ and $L$ candidate activity labels, multi-label classification is to predict $\{y_l\}_{1\leq l \leq L}$, where $y_l \in \{0, 1\}$ is a binary indicator to describe whether the class $l$ is present in the sequence. A straightforward multi-label classification method is Binary Relevance (BN): the features extracted by the BiGRU are averaged and then fed into $L$ independent binary fully connected classifiers to predict $\{y_l\}_{1\leq l \leq L}$. This model is denoted as BiGRU+BN. To enhance the model's ability to extract label correlations and pay attention to important features of the sequence, we further propose the BiGRU+Q2L model, depicted in Figure~\ref{fig:Q2L}, which utilizes the Query2Label~(Q2L)~\cite{liu2021query2label} model as the multi-label classifier.

Query2Label embeds candidate labels into a label embedding vector and then feeds them into a Transformer decoder, each layer consisting of a self-attention module, a cross-attention module, and a position-wise feed-forward network. In the self-attention module, query, key, and value are all the label embedding vectors. Unlike binary relevance, the correlation between labels can be learned in this module. In the cross-attention module, the queries are label embeddings, whereas keys and values are the temporal features extracted by the encoder. This module allows each label to be associated with its desired features and pool them by linear combinations. The queried feature vectors are then fed to the position-wise feed-forward networks for further non-linear transformations.

Therefore, the output vectors at each label position in the Transformer decoder are a fusion and transformation of the label-related features of the input sequence. We apply a linear transformation to each output, followed by the sigmoid function to frame it into the range $[0,1]$. This numerical value, denoted as $p_l$, represents the probability of presence of the label $l$. Empirically, labels with probabilities greater than $0.7$ are finally considered as output predictions. Binary Cross Entropy Loss (BCE Loss) is used as the loss function.

\subsection{Multi-label classification with resident separation}
\label{HAR_s2s_sep}

% \todo{Explain how each model (BiGRU+BN and BiGRU+Q2L) can be adapted to use separated sequences}
As mentioned earlier, the distinction between resident separation and multi-label classification lies in the timing of information separation: they are not mutually exclusive. The proposed Seq2Res and BiGRU+Q2L/BN models can be combined into a two-stage model, where Seq2Res is first used to perform resident separation on the mixed sequence, and then the output separated sequence is treated as a whole input for activity recognition in BiGRU+Q2L/BN. Compared to directly conducting individual activity recognition after resident separation, this two-stage approach allows for co-consideration of sensor events from both individuals before classification. The input of BiGRU+Q2L is the sequence of softmax probability vectors from the output of the fully connected network of Seq2Res, rather than the exact most probable events, to retain the most information.

% Combining the previously discussed methods for resident separation and multi-resident activity recognition, the following approach is proposed. The original event sequence is first separated by the Seq2Seq resident separation model and then input into the Query2Label classification model for multi-label classification. For the output of Seq2Seq, we do not generate the final separation sequence. Instead, we use the softmax probabilities vector of each event outputted by the decoder for the purpose of maximizing the retention of information.

\section{Experimental Results}
\label{section:exps}

%In this section, we present the CASAS ADLMR dataset~\cite{singla2010recognizing}, describe our experimental setup, and provide the corresponding evaluation results for both resident separation and multi-person activity recognition.

\subsection{Dataset}
The proposed approaches are evaluated on a real-world dataset known as the Multiresident ADL Activities (ADLMR) dataset\footnote{ \url{http://casas.wsu.edu/datasets/adlmr.zip} (last seen on 11/2023)}, which was published by the Center for Advanced Studies in Adaptive Systems (CASAS) of the Washington State University \cite{singla2010recognizing}. This dataset comprises 26 subsets, each representing a single day, with each day containing sensor events triggered by 2 residents (a different pair each day) performing activities among a shared set of 15 classes. There is a total of 37 different ambient sensors in the home, such as motion and opening sensors. Each sensor event has been manually labeled with the identity of the resident triggering this event, along with the activity they were engaged in. Therefore, this dataset can be used for both resident separation as in~\cite{riboni2020unsupervised} and multi-subject activity recognition as in~\cite{chen2022transformer}.

\subsection{Experimental setup}
% Todo: Experimental setup to compare performance.

\subsubsection{Evaluation method} For each experiment, we used 10-fold cross-validation. Data was partitioned using the scikit-learn library. Since each day is performed by different pairs of residents, one day can not be split into different folds to ensure cross-resident independence. As such, each fold contains entire days. In order to cover the whole 26 days, the test set in the first 6 folds contains data for 3 days, ranging from day 1 to day 18; in the subsequent 4 folds, each test set contains data for 2 days, covering days 18 to 26.

\subsubsection{Data preparation} Due to typographical errors in the annotations of the original dataset, we corrected some of the labels. In addition, for events with labels of single-resident activity, we assigned the last performed activity of the other resident as a second label, so that each event is labeled with the activity of both residents. To reduce data redundancy, we excluded events in which motion sensors were automatically deactivated. By applying a sliding window approach, we segmented data so that each data instance contains 16 sensor events, with a step of 3 events between each instance. The activity label of an instance is the result of majority voting between the labels of the last 3 events.

\subsubsection{Parameters} All parameters were set following preliminary experiments. For the encoder of Seq2Res, the embedding size and the hidden size of BiGRU are $128$. The output size and the context vector size are then $2\times128 = 256$. A dropout rate of $0.1$ is used. For the decoder, the embedding size and hidden size are both $256$. Given that the input to the GRU is a concatenation of the embedded input vector and the attention-queried vector, the input size for GRU is $2\times256 = 512$. A dropout rate of $0.4$ is used for the decoder. The initial learning rate for training is $0.001$, with a halving schedule every $80$ epochs. Seq2Res was trained for a total of $300$ epochs.

For BiGRU+Q2L, the event embedding size and the hidden size of BiGRU are both $128$, resulting in an output feature vector size of $2\times128 = 256$. The label embedding size of Q2L is consequently set to $256$. A dropout rate of $0.3$ is used. A learning rate of $1\times10^{-4}$ is used for a $100$ epochs training. The Adam optimizer is used, and a batch size of $100$ is used across all training sessions.

\subsection{Metrics}
\subsubsection{Resident separation}

\begin{figure}
    \centering
    \includegraphics[width=1.0\linewidth]{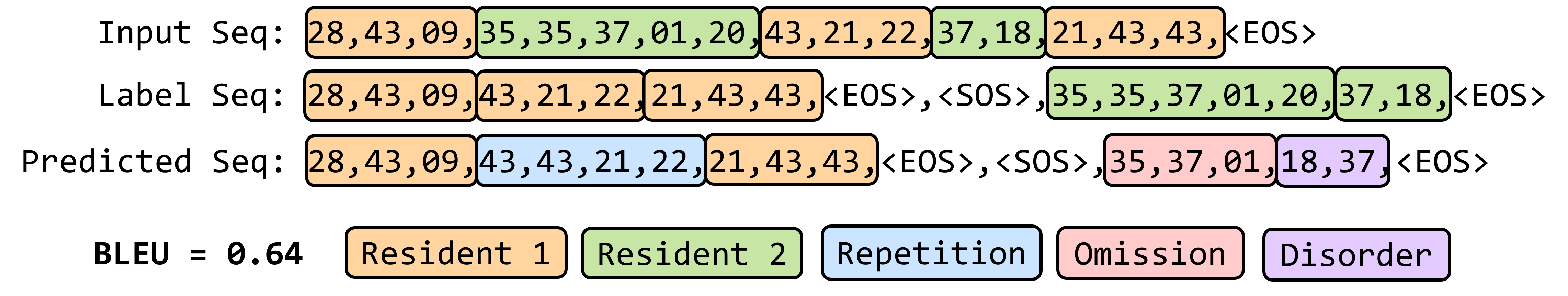}
    \caption{Example of an input sequence of Seq2Res model with its label and prediction. Each number represents the token of a sensor event.}
    %\caption{Example of an input sequence of Seq2Res model with its label and prediction. Each number represents the token of a sensor event. The BLEU~\cite{papineni2002bleu} score between prediction and label is 0.64. Three error types (repetition, omission, and disorder) are color-marked.}
    \label{fig:bleu_example}
\vspace{-1em}
\end{figure}

Former resident separation compute typically the accuracy of each event assignment through one-to-one comparisons between the prediction and the ground truth in terms of their positions. Our separation method consists of sequence generation, making it challenging to establish a direct one-to-one alignment. Moreover, calculating accuracy on a one-to-one alignment fails to consider the coherence of the separated sequence, which are crucial for extracting temporal information from the sequence. Hence, we borrowed a metric commonly used in machine translation, Bilingual Evaluation Understudy~(BLEU)~\cite{papineni2002bleu}, to assess our separation results. An illustrative example is given in Figure~\ref{fig:bleu_example}, in which the BLEU score between the prediction and the label is 0.64.

Given a generated sequence $c$ and a reference sequence $r$, the BLEU metric considers the precision of $N$-grams, which is the proportion of $N$-gram phrases in the generated sequence $c$ that appear in the reference sequence $r$, and penalizes shorter sequences. Like most studies, the final BLEU of this research is the average of scores corresponding to $N$ from 1 to 4.

\subsubsection{Activity Recognition}
We use the standard metrics of accuracy, recall, precision, and F1 score.

\subsection{Results on resident separation}
\begin{table}[t]
\centering
\caption{Performance of resident separation with Seq2Res.}
\resizebox{0.49\textwidth}{!}{
\begin{tabular}{|cccc|}
\hline
\multicolumn{2}{|c|}{BLEU}                                                                     & \multicolumn{1}{c|}{Seq2Res (ours)}   & SMRT \cite{wang2020smrt} \\ \hline
\multicolumn{1}{|c|}{\multirow{15}{*}{Class}} & \multicolumn{1}{c|}{Fill medication dispenser}  & \multicolumn{1}{c|}{\textbf{0.7906}} & 0.6152  \\
\multicolumn{1}{|c|}{}                          & \multicolumn{1}{c|}{Hang up clothes}            & \multicolumn{1}{c|}{\textbf{0.8397}} & 0.7225  \\
\multicolumn{1}{|c|}{}                          & \multicolumn{1}{c|}{Move couch and table}       & \multicolumn{1}{c|}{\textbf{0.4844}} & 0.4326  \\
\multicolumn{1}{|c|}{}                          & \multicolumn{1}{c|}{Read on couch (user B)}     & \multicolumn{1}{c|}{\textbf{0.4799}} & 0.4509  \\
\multicolumn{1}{|c|}{}                          & \multicolumn{1}{c|}{Water plants}               & \multicolumn{1}{c|}{0.7174} & \textbf{0.7411}  \\
\multicolumn{1}{|c|}{}                          & \multicolumn{1}{c|}{Sweep kitchen floor}        & \multicolumn{1}{c|}{\textbf{0.6212}} & 0.4899  \\
\multicolumn{1}{|c|}{}                          & \multicolumn{1}{c|}{Play checkers}              & \multicolumn{1}{c|}{\textbf{0.6389}} & 0.5016  \\
\multicolumn{1}{|c|}{}                          & \multicolumn{1}{c|}{Set out dinner ingredients} & \multicolumn{1}{c|}{\textbf{0.7017}} & 0.6387  \\
\multicolumn{1}{|c|}{}                          & \multicolumn{1}{c|}{Set dinner table}           & \multicolumn{1}{c|}{\textbf{0.6993}} & 0.6538  \\
\multicolumn{1}{|c|}{}                          & \multicolumn{1}{c|}{Read on couch (user A)}     & \multicolumn{1}{c|}{\textbf{0.5713}} & 0.5372  \\
\multicolumn{1}{|c|}{}                          & \multicolumn{1}{c|}{Pay electric bill}          & \multicolumn{1}{c|}{\textbf{0.5619}} & 0.5445  \\
\multicolumn{1}{|c|}{}                          & \multicolumn{1}{c|}{Prepare picnic basket}      & \multicolumn{1}{c|}{\textbf{0.6086}} & 0.6073  \\
\multicolumn{1}{|c|}{}                          & \multicolumn{1}{c|}{Retrieve dishes}            & \multicolumn{1}{c|}{\textbf{0.5887}} & 0.5513  \\
\multicolumn{1}{|c|}{}                          & \multicolumn{1}{c|}{Pack supplies in basket}    & \multicolumn{1}{c|}{\textbf{0.6290}} & 0.4708  \\
\multicolumn{1}{|c|}{}                          & \multicolumn{1}{c|}{Pack food in basket}        & \multicolumn{1}{c|}{\textbf{0.6266}} & 0.4864  \\ \hline
\multicolumn{2}{|c|}{Overall BLEU}                                                                     & \multicolumn{1}{c|}{\textbf{0.6385}}   & 0.5608 \\ \hline
% \multicolumn{4}{|c|}{Overall BLEU of Seq2Res: \textbf{0.6385}} \\ \hline
% \multicolumn{4}{|c|}{Overall BLEU of SMRT \cite{wang2020smrt}: 0.5608}\\ \hline
\end{tabular}
}
\vspace{-1em}
\label{tab: sep_bleu}
\end{table}

We conduct experiments on resident separation using the Seq2Res model and reproduce the SMRT (Sensor-based Multi-resident Tracking) \cite{wang2020smrt} under the same protocol for comparison. Table~\ref{tab: sep_bleu} reports, for both methods, the cross-validation average BLEU for each activity class and the overall average BLEU. We see that the overall performance of Seq2Res is higher than that of SMRT. This could be attributed to the fact that Seq2Rees, compared to SMRT, is better able to consider a longer context (by the encoder) while ensuring the correlation between two consecutive sensor events (by the decoder). The overall BLEU of Seq2Res reaches $0.6385$. Three types of error are generally observed in generated sequences: repetition, omission and disorder, as illustrated in Figure~\ref{fig:bleu_example}. A significant variation across different classes is also observed. For example, ``\textit{Move couch and table}'' and ``\textit{Read on couch (user B)}'' have BLEU scores below $0.5$. These two activities always occur in close proximity, and the trajectories of the two residents significantly overlap, making it more difficult to separate. Conversely, ``\textit{Fill medication dispense}'' and ``\textit{Hang up clothes}'' take place at opposite sides of the house, with minimal overlap in the trajectories of the residents. As a result, the BLEU scores for these two activities reach around $0.8$. In general, the separation ability of the model is negatively correlated with the degree of overlap in the actions of the two residents, which conforms to intuition.

% It is important to note that a sequence often corresponds to 1 or 2 classes. The BLEU of each sequence is assigned to all classes in that sequence.
% We calculated the BLEU between the separated sequence and the ground truth sequence and then allocated it to the corresponding categories of the sequence.

% Figure X summarises the results of resident separation for each fold. 
% We observed that our model is capable of generating relatively accurate separated subsequences, and further improvements in results are attainable through the application of data augmentation.
% Nevertheless, it is important to note that there are noticeable variations in the evaluation results among different folds. This variance can be attributed to the difference of residents in each fold, which subsequently decides their different behavior patterns and, consequently, the variable performance.

\subsection{Results on activity recognition} \label{sec: HAR_results}
In this section, 3 scenarios are used to compare the performance of multi-resident activity recognition models:
\begin{itemize}
    \item No\_Sep: Inputs are event sequences without separation.
    \item S2S\_Sep: Inputs are event sequences generated by Seq2Res as introduced in Section~\ref{HAR_s2s_sep}.
    \item GT\_Sep: Inputs are event sequences separated based on the ground truth labeled in the dataset.
    
\end{itemize}

% The only difference between these three scenarios lies in their inputs: Given a sensor event sequence $\{e_k\}$ that can be separated into $\{e_k^1\}$ and $\{e_k^2\}$, The input of A is the unseparated sequence $\{e_k\}$, while the input of B is the separated sequence predicted by a separation model $Sep(\{e_k\})$, and in scenario C, the input is the separated sequence accurately separated based on the data labels, $\{\{\Tilde{e}_k^1\}, <EOS>, <SOS>, \{\Tilde{e}_k^2\}\}$, which are also the label sequences trained by the separation model for B.

Under these 3 scenarios, we first evaluate the recognition accuracy and macro-F1 score of 3 different models: TransBiGRU\cite{chen2022transformer} using a binary relevance classifier, BiGRU+BN and BiGRU+Q2L (both presented in Section~\ref{section:q2l}). For TransBiGRU, we used the same hyperparameters as in \cite{chen2022transformer}.

\begin{table}[t]
\centering
\caption{Performance of activity recognition models for 3 scenarios.}
\resizebox{0.49\textwidth}{!}{
\begin{tabular}{|c|c|c|c|}
\hline
Scenario                  & Model              & Accuracy (\%) & Macro-F1 (\%) \\ \hline
\multirow{3}{*}{No\_Sep}  & BiGRU+BN              & 87.66 (0.31)  & 86.09 (0.37)  \\
                          & TransBiGRU+BN\cite{chen2022transformer} & 86.85 (0.34)              & 85.13 (0.38)             \\
                          & \textbf{BiGRU+Q2L (ours)}         & \textbf{88.47 (0.23)}  & \textbf{87.07 (0.25)}  \\ \hline
\multirow{3}{*}{S2S\_Sep} & BiGRU+BN              &  79.36 (0.54)  & 76.74 (0.70)  \\
                          & TransBiGRU+BN\citet{chen2022transformer} & 73.38 (0.56)              & 70.12 (0.66)              \\
                          & \textbf{BiGRU+Q2L (ours)}          & \textbf{83.26 (0.39)}  & \textbf{81.08 (0.47)}  \\ \hline
\multirow{3}{*}{GT\_Sep}  & BiGRU+BN              & 88.70 (0.40)  & 87.15 (0.48)  \\
                          & TransBiGRU+BN\citet{chen2022transformer} & 87.35 (0.38)             & 85.68 (0.44)             \\
                          & \textbf{BiGRU+Q2L (ours)}         & \textbf{90.87 (0.32)}  & \textbf{89.57 (0.38)}  \\ \hline
\end{tabular}
}
\vspace{-1em}
\label{tab: HAR_acc_F1}
\end{table}
\begin{table*}[t]
\centering
% \begin{table*}[htbp]
\caption{Performance of BiGRU+Q2L for each activity class, for 3 scenarios.}
\resizebox{0.99\textwidth}{!}{
\begin{tabular}{|cc|ccc|ccc|ccc|c|}
\hline
\multicolumn{2}{|c|}{Metric}                                       & \multicolumn{3}{c|}{Precision (\%)}                                                       & \multicolumn{3}{c|}{Recall (\%)}                                       & \multicolumn{3}{c|}{F1-score (\%)}                                                        & \multirow{2}{*}{Count} \\ \cline{1-11}
\multicolumn{2}{|c|}{Scenario}                                     & \multicolumn{1}{c|}{No\_Sep} & \multicolumn{1}{c|}{S2S\_Sep} & GT\_Sep                    & \multicolumn{1}{c|}{No\_Sep} & \multicolumn{1}{c|}{S2S\_Sep} & GT\_Sep & \multicolumn{1}{c|}{No\_Sep} & \multicolumn{1}{c|}{S2S\_Sep} & GT\_Sep                    &                        \\ \hline
\multicolumn{1}{|c|}{\multirow{15}{*}{Class}} & Fill medication dispenser  & \multicolumn{1}{c|}{89.54}   & \multicolumn{1}{c|}{85.79}    & 93.94                      & \multicolumn{1}{c|}{92.71}   & \multicolumn{1}{c|}{87.55}    & 96.12   & \multicolumn{1}{c|}{91.10}   & \multicolumn{1}{c|}{86.66}    & 95.02                      & 4770                   \\
\multicolumn{1}{|c|}{}                          & Hang up clothes   & \multicolumn{1}{c|}{89.97}   & \multicolumn{1}{c|}{85.19}    & 88.44                      & \multicolumn{1}{c|}{92.13}   & \multicolumn{1}{c|}{87.65}    & 90.51   & \multicolumn{1}{c|}{91.04}   & \multicolumn{1}{c|}{86.40}    & 89.46                      & 2890                   \\
\multicolumn{1}{|c|}{}                          & Move couch and table    & \multicolumn{1}{c|}{86.00}   & \multicolumn{1}{c|}{75.92}    & 87.80                      & \multicolumn{1}{c|}{84.01}   & \multicolumn{1}{c|}{79.79}    & 86.63   & \multicolumn{1}{c|}{84.99}   & \multicolumn{1}{c|}{77.81}    & 87.21                      & 2450                   \\
\multicolumn{1}{|c|}{}                          & Read on couch (user B) & \multicolumn{1}{c|}{84.15}   & \multicolumn{1}{c|}{77.43}    & 90.58                      & \multicolumn{1}{c|}{84.22}   & \multicolumn{1}{c|}{80.92}    & 90.92   & \multicolumn{1}{c|}{84.18}   & \multicolumn{1}{c|}{79.13}    & 90.75                      & 2410                   \\
\multicolumn{1}{|c|}{}                          & Water plants      & \multicolumn{1}{c|}{82.25}   & \multicolumn{1}{c|}{76.10}    & 86.10                      & \multicolumn{1}{c|}{85.06}   & \multicolumn{1}{c|}{78.62}    & 88.72   & \multicolumn{1}{c|}{83.63}   & \multicolumn{1}{c|}{77.33}    & 87.39                      & 2000                   \\
\multicolumn{1}{|c|}{}                          & Sweep kitchen floor & \multicolumn{1}{c|}{92.58}   & \multicolumn{1}{c|}{87.81}    & 93.90                      & \multicolumn{1}{c|}{89.26}   & \multicolumn{1}{c|}{85.61}    & 93.04   & \multicolumn{1}{c|}{90.89}   & \multicolumn{1}{c|}{86.70}    & 93.47                      & 4840                   \\
\multicolumn{1}{|c|}{}                          & Play checkers     & \multicolumn{1}{c|}{93.64}   & \multicolumn{1}{c|}{90.64}    & 95.53                      & \multicolumn{1}{c|}{90.94}   & \multicolumn{1}{c|}{87.79}    & 93.15   & \multicolumn{1}{c|}{92.27}   & \multicolumn{1}{c|}{89.19}    & 94.33                      & 5910                   \\
\multicolumn{1}{|c|}{}                          & Set out dinner ingredients   & \multicolumn{1}{c|}{80.05}   & \multicolumn{1}{c|}{73.96}    & 83.30                      & \multicolumn{1}{c|}{80.95}   & \multicolumn{1}{c|}{77.29}    & 89.92   & \multicolumn{1}{c|}{80.50}   & \multicolumn{1}{c|}{75.59}    & 86.48                      & 1970                   \\
\multicolumn{1}{|c|}{}                          & Set dinner table         & \multicolumn{1}{c|}{83.71}   & \multicolumn{1}{c|}{79.77}    & 86.41                      & \multicolumn{1}{c|}{83.56}   & \multicolumn{1}{c|}{80.51}    & 86.33   & \multicolumn{1}{c|}{83.63}   & \multicolumn{1}{c|}{80.14}    & 86.37                      & 3480                   \\
\multicolumn{1}{|c|}{}                          & Read on couch (user A) & \multicolumn{1}{c|}{81.08}   & \multicolumn{1}{c|}{73.77}    & 83.70                      & \multicolumn{1}{c|}{79.34}   & \multicolumn{1}{c|}{75.16}    & 81.86   & \multicolumn{1}{c|}{80.20}   & \multicolumn{1}{c|}{74.46}    & 82.77                      & 2970                   \\
\multicolumn{1}{|c|}{}                          & Pay electric bill        & \multicolumn{1}{c|}{84.72}   & \multicolumn{1}{c|}{79.06}    & 86.22                      & \multicolumn{1}{c|}{82.39}   & \multicolumn{1}{c|}{77.99}    & 83.06   & \multicolumn{1}{c|}{83.54}   & \multicolumn{1}{c|}{78.52}    & 84.61                      & 3070                   \\
\multicolumn{1}{|c|}{}                          & Prepare picnic basket   & \multicolumn{1}{c|}{91.05}   & \multicolumn{1}{c|}{91.12}    & 94.83                      & \multicolumn{1}{c|}{91.42}   & \multicolumn{1}{c|}{85.96}    & 93.62   & \multicolumn{1}{c|}{91.23}   & \multicolumn{1}{c|}{88.46}    & 94.22                      & 5710                   \\
\multicolumn{1}{|c|}{}                          & Retrieve dishes   & \multicolumn{1}{c|}{90.02}   & \multicolumn{1}{c|}{90.97}    & 93.08                      & \multicolumn{1}{c|}{92.35}   & \multicolumn{1}{c|}{86.59}    & 93.05   & \multicolumn{1}{c|}{91.17}   & \multicolumn{1}{c|}{88.73}    & 93.06                      & 5750                   \\
\multicolumn{1}{|c|}{}                          & Pack supplies in basket    & \multicolumn{1}{c|}{83.44}   & \multicolumn{1}{c|}{70.13}    & 85.10                      & \multicolumn{1}{c|}{85.18}   & \multicolumn{1}{c|}{74.48}    & 87.13   & \multicolumn{1}{c|}{84.30}   & \multicolumn{1}{c|}{72.24}    & 86.19                      & 3080                   \\
\multicolumn{1}{|c|}{}                          & Pack food in basket         & \multicolumn{1}{c|}{84.38}   & \multicolumn{1}{c|}{69.84}    & \multicolumn{1}{c|}{84.25} & \multicolumn{1}{c|}{84.88}   & \multicolumn{1}{c|}{73.88}    & 85.13   & \multicolumn{1}{c|}{84.63}   & \multicolumn{1}{c|}{71.81}    & \multicolumn{1}{c|}{84.69} & 3220                   \\ \hline
\multicolumn{2}{|c|}{Average}                                       & \multicolumn{1}{c|}{86.44}   & \multicolumn{1}{c|}{80.50}    & \multicolumn{1}{c|}{88.88} & \multicolumn{1}{c|}{86.56}   & \multicolumn{1}{c|}{81.32}    & 89.28   & \multicolumn{1}{c|}{86.49}   & \multicolumn{1}{c|}{80.88}    & \multicolumn{1}{c|}{89.06} & 3635                   \\ \hline
\end{tabular}
}
\vspace{-1em}
\label{tab: HAR_metrics_per_class}
\end{table*}
Table~\ref{tab: HAR_acc_F1} reports the average cross-validation performance of these 3 models for each scenario, with standard deviations in parentheses. BiGRU+Q2L achieves the best performance for all 3 scenarios. Compared to the BiGRU+BN model, BiGRU+Q2L exhibits statistically significant improvement in performances, especially for macro-F1 compared to accuracy. This indicates that the Query2Label classifier can help address data imbalance between activity classes, which is a common problem in human activity recognition. The performance of TransBiGRU~\cite{chen2022transformer} is significantly lower than BiGRU+BN, a lighter model with fewer parameters, for all scenarios. For the S2S\_Sep scenario, where the difference is significantly large, the excessively deep network of TransBiGRU is overfitting on the noise in the generated separated sequences, resulting in significant drops in performance.

Comparing the results across the three scenarios, we observe that GT\_Sep (using ground truth resident labels) has an accuracy and macro-F1 score that are $2.4\%$ and $2.5\%$ higher, respectively, than No\_Sep in the BiGRU+Q2L model (and similar gaps for the other 2 models). This indicates that resident separation does help multi-resident activity recognition when this separation is perfectly accurate. Performances for the S2S\_Sep scenario reach the same orders of magnitude, with accuracies as high as $83.26\%$ with BiGRU+Q2L. However, they are significantly lower than for both GT\_Sep and No\_Sep scenarios, for all models. This suggests that, although the separated sequences generated by the Seq2Res model are overall representative of the true separation, the errors introduced during the generation process have a significant impact on the final activity classification. In general, these results show that resident separation can improve activity recognition, but only if the separated sequences are very accurate, which remains a scientific challenge.

To further investigate the behavior of models for each scenario, we report in Table~\ref{tab: HAR_metrics_per_class} the cross-validation average precision, recall, and F1 scores of the BiGRU+Q2L model, per activity class. We observe that models for the S2S\_Sep scenario achieve better results for classes with larger numbers of instances (e.g. ``\textit{Play checkers}'', ``\textit{Prepare picnic basket}'', ``\textit{Retrieve dishes}''), but underperform for classes with a smaller number of instances (e.g. ``\textit{Water plants}'', ``\textit{Set out dinner ingredients}''). This could be because the noise introduced by generative resident separation increases the variation of the classifier input, which requires more instances to learn. On the other hand, comparing GT\_Sep and No\_Sep highlights that accurate resident separation helps to recognize classes with a small number of instances (e.g. ``\textit{Set out dinner ingredients}'', ``\textit{Read on couch (user B)}'') because the temporal features of the accurately separated sequences may be easier to learn.

% The results of the evaluation with respect to the 3 scenarios are listed in Table \ref{tab: HAR_metrics_per_class}. We observed that in scenario C, the accuracy is higher than in scenario A by $2$ to $3\%$ for most categories, and it is even 5 percentage points higher for categories 12 and 13. This indicates that accurate user separation does indeed improve activity recognition accuracy. However, scenario B underperforms, with lower accuracy compared to scenario A in most categories, but it outperforms scenario A in categories 12 and 13, which rely more on resident separation.

% Table \ref{tab: HAR_metrics_per_class} presents respectively the recognition results for these three configurations. We observed that using ground truth for separation led to a notable $3\%$ increase in recognition accuracy when compared to the unseparated scenario. This underscores the potential utility of resident separation methods in enhancing recognition performance. However, using our model for separation did not lead to an improvement in overall activity recognition performance compared to the unseparated scenario.

% Figure X further illustrates the recognition performance for each category in the scenarios. By comparing the non-separation and ground-truth-based separation approaches, we found that ground-truth-based separation improved the model's accuracy in recognizing the 12th and 13th classes. 

\section{Conclusion}
\label{section:conclusion}

In this paper, two models are presented: Seq2Res for resident separation, and BiGRU+Q2L for multi-resident activity recognition. On the CASAS ADLMR dataset, BiGRU+Q2L achieves better performance than another state-of-the-art model TransBiGRU, with a simpler architecture. While the Seq2Res model shows potential for the resident separation task, the quality of the generated sequences is still limited. As such, the combination of Seq2Res and BiGRU+Q2L does not yet reach the same performance as using only BiGRU+Q2L.

Experiments with ground truth separation have highlighted that using perfect resident separation before multi-resident activity recognition can significantly improve performance. Therefore, future work on improving resident separation must be conducted, such as post-processing methods to improve the sequences generated by Seq2Res.

\bibliographystyle{plainnat}
\bibliography{refs}

\end{document}